\documentclass[sigconf]{acmart}

\def\BibTeX{{\rm B\kern-.05em{\sc i\kern-.025em b}\kern-.08emT\kern-.1667em\lower.7ex\hbox{E}\kern-.125emX}}

\usepackage{amsmath}
\usepackage{array}
\usepackage{graphicx}
\usepackage{subfigure}

\usepackage{multirow}

\usepackage{amssymb}

\usepackage{xspace}
\makeatletter
\DeclareRobustCommand\onedot{\futurelet\@let@token\@onedot}
\def\@onedot{\ifx\@let@token.\else.\null\fi\xspace}

\def\eg{\emph{e.g}\onedot} 
\def\ie{\emph{i.e}\onedot}

\makeatother

\usepackage{lipsum}
\makeatletter
\def\@IEEEsectpunct{.\ \,}
\def\paragraph{\@startsection{paragraph}{4}{\z@}{1.5ex plus 1.5ex minus 0.5ex}%
	{0ex}{\normalfont\normalsize\sffamily\bfseries}}
\makeatother

\newcommand{\bftab}{\fontseries{b}\selectfont}

\newcommand\blfootnote[1]{%
	\begingroup
	\renewcommand\thefootnote{}\footnote{#1}%
	\addtocounter{footnote}{-1}%
	\endgroup
}

\copyrightyear{2019} 
\acmYear{2019} 
\acmConference[MM '19]{Proceedings of the 27th ACM International Conference on Multimedia}{October 21--25, 2019}{Nice, France}
\acmBooktitle{Proceedings of the 27th ACM International Conference on Multimedia (MM '19), October 21--25, 2019, Nice, France}
\acmPrice{15.00}
\acmDOI{10.1145/3343031.3350943}
\acmISBN{978-1-4503-6889-6/19/10}

\begin{document}

\fancyhead{}

\title[Aligning Linguistic Words and Visual Semantic Units for Image Captioning]{Aligning Linguistic Words and Visual Semantic Units \\for Image Captioning}

\author{Longteng Guo$^{1,4}$, Jing Liu$^{1*}$, Jinhui Tang$^{2}$, Jiangwei Li$^{3}$, Wei Luo$^{3}$, Hanqing Lu$^{1}$} 
\affiliation{\small
	$^1$National Laboratory of Pattern Recognition, Institute of Automation, Chinese Academy of Sciences\; \\
	$^2$School of Computer Science and Engineering, Nanjing University of Science and Technology\; \\
	$^3$Multimedia Department, Huawei Devices\; 
	$^4$University of Chinese Academy of Sciences\; 
}
\affiliation{ \small \{longteng.guo,jliu,luhq\}@nlpr.ia.ac.cn\; jinhuitang@njust.edu.cn\; \{lijiangwei1,luo.luowei\}@huawei.com}

\renewcommand{\shortauthors}{Longteng Guo et al.}

\begin{abstract}
Image captioning attempts to generate a sentence composed of several linguistic words, which are used to describe objects, attributes, and interactions in an image, denoted as visual semantic units in this paper. Based on this view, we propose to explicitly model the object interactions in semantics and geometry based on Graph Convolutional Networks (GCNs), and fully exploit the alignment between linguistic words and visual semantic units for image captioning. Particularly, we construct a semantic graph and a geometry graph, where each node corresponds to a visual semantic unit, i.e., an object, an attribute, or a semantic (geometrical) interaction between two objects. Accordingly, the semantic (geometrical) context-aware embeddings for each unit are obtained through the corresponding GCN learning processers. At each time step, a context gated attention module takes as inputs the embeddings of the visual semantic units and hierarchically align the current word with these units by first deciding which type of visual semantic unit (object, attribute, or interaction) the current word is about, and then finding the most correlated visual semantic units under this type. Extensive experiments are conducted on the challenging MS-COCO image captioning dataset, and superior results are reported when comparing to state-of-the-art approaches. The code is publicly available at \url{https://github.com/ltguo19/VSUA-Captioning}.
\end{abstract}

\ccsdesc[500]{Computing methodologies~Natural language generation}
\ccsdesc[300]{Computing methodologies~Computer vision}
\ccsdesc[100]{Computing methodologies~Image representations}

\keywords{image captioning; graph convolutional networks; visual relationship; visual-language}

\maketitle

\section{Introduction}
\blfootnote{* Corresponding Author}Computer vision and natural language processing are becoming increasingly intertwined. 
At the intersection of the two subjects, automatically generating lingual descriptions of images, namely image captioning \cite{vinyals2017show,karpathy2015deep,Fang2014From}, has emerged as a prominent interdisciplinary research problem.
Modern image captioning models typically employ an encoder-decoder framework, where the encoder encodes an image into visual representations and then the decoder decodes them into a sequence of words. %

\begin{figure}[!t] 
	\centering
	\includegraphics[width=3.3in]{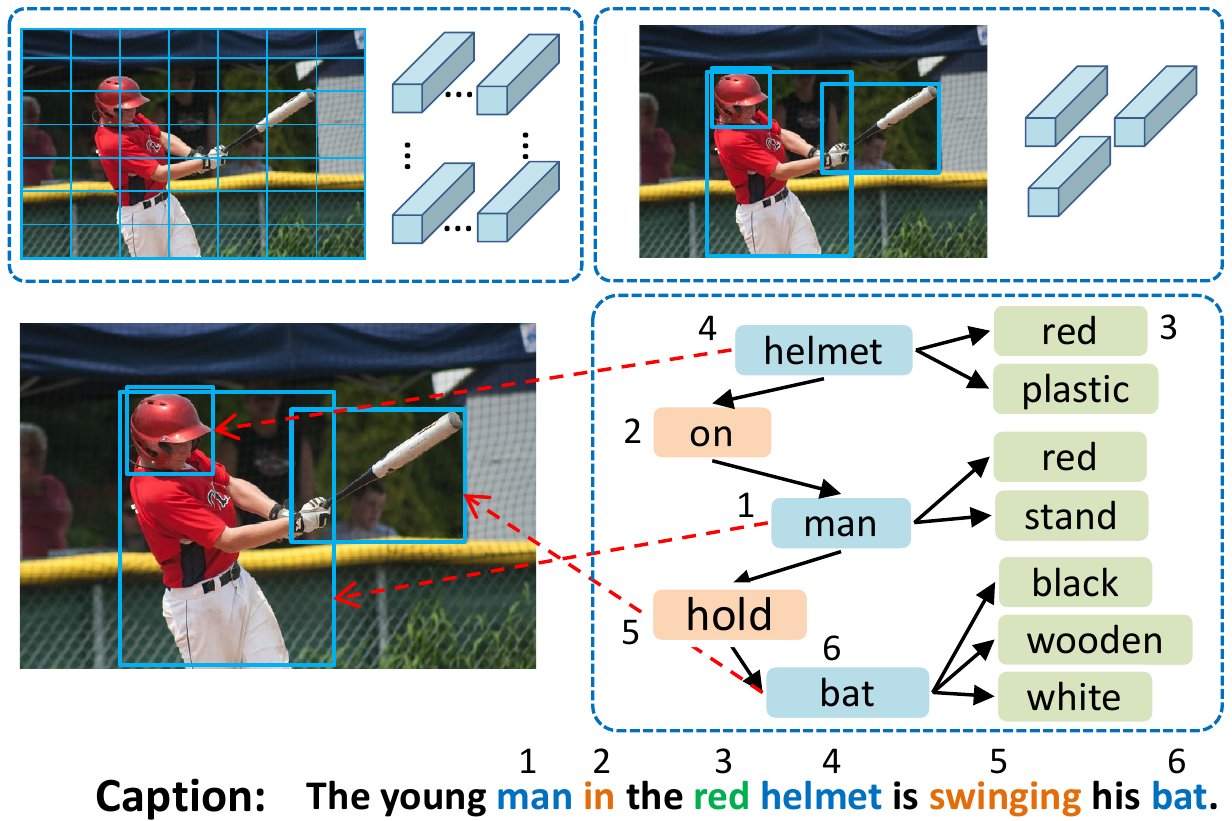}
	\caption{
		Typically, image captioning models consider the visual content of images as uniform grids (top left) or collections of object regions (top right).
		Differently, our approach represents images as structured graphs where nodes are VSUs: object, attribute, and relationship units (bottom).
		We make use of the alignment nature between caption words and VSUs. 
	}
	\vspace{-0.3cm}
	\label{fig:semantic_units}
\end{figure}

How to represent visual content and how to reason over them are fundamental problems in image captioning. 
Starting from the static, single-vector representations, the visual representations have evolved into using dynamic, multi-vector representations, which are often fed into an attention module for information aggregation. 
In the early times, the image is treated as uniform grid representations, then more recently, state-of-the-art methods regard visual content as collections of individual object regions (the top row in Figure~\ref{fig:semantic_units}). 
However, the isolated objects only represent the categories and properties of individual instances, which are often related to nouns or noun phrases in a caption, but fail to model object-object relations, \eg the interactions or relative positions. While the relationships between objects are the natural basis for describing an image.

In fact, the image is a structured combination of objects (``man", ``helmet"), their attributes (``helmet is red"), and more importantly, relationships (``man hold bat") involving these objects. 
We call these visual components \textbf{visual semantic units} (\textbf{VSUs}) in this paper, which include three categories: object units, attribute units, and relationship units.  
At the same time, a sentence is also composed of syntactic units describing objects (\eg nouns phrase), their properties (\eg adjectives) and relations  (\eg verb, prepositions). 
Because captions are abstractions of images, it is intuitive that each word in the caption can roughly be aligned with the VSUs of the image. 
Exploiting such vision-language correlation could benefit image understanding and captioning.

In this paper, we propose to represent visual content with VSU-based structured graphs,   
and take advantage of the strong correlations between linguistic syntactic units and VSUs for image captioning.
First, we detect a collection of VSUs from the image through object, attribute, and relationship detectors, respectively. 
Then we construct \textit{structured graphs} as explicit and unified representation that connects the detected objects, attributes, and relationships  (\eg bottom-right in Figure~\ref{fig:semantic_units}), 
where each node corresponds to a VSU and the edges are the connections between two VSUs.  
In particular, we construct a semantic graph and a geometry graph, where the former models the semantic relationships (\eg ``man holding bat") 
and the latter models the geometry relationships (\eg ``the man the and bat are adjacent and overlapped"). 
After that, Graph Convolutional Networks (GCNs) are then explored to learn context-aware embeddings for each VSU in the graph. %

We design a context gated attention module (CGA) that attends to the three types of VSUs in a \textit{hierarchical} manner when generating each caption word. 
The key insight behind CGA is that each word in the caption could be aligned with a VSU, 
and if the word is about objects, \eg a noun, then the corresponding VSU should also be an object unit, 
meaning that more attention should be paid on the object units. 
Specifically, CGA first performs three independent attention mechanism inside the object, attribute, and relationship units, respectively. 
Then a gated fusion process is performed to adaptively decide how much attention should be paid to each of the three VSU categories by referring to the current linguistic context. Knowledge learned from the semantic graph and the geometry graph are naturally fused by extending CGA's input components to include the VSUs from both graphs. 

The main contributions of this paper are three-fold.
\begin{itemize}
	\item We introduce visual semantic units as comprehensive representation of the visual content, 
	and exploit structured graphs, \ie semantic graph and geometry graph, and GCNs to uniformly represent them. 
	\item We explore the vision-language correlation and design a context gated attention module to hierarchically align linguistic words and visual semantic units. 
	\item Extensive experiments on MS COCO validates the superiority of our method. 
	Particularly, in terms of the popular CIDEr-D metric, we achieve an absolute $8.5$ points improvement over the strong baseline, \ie Up-Down \cite{anderson2017bottom}, on Karpathy test split. 
\end{itemize}

\section{Related Work}
\paragraph{Image Captioning. }
Filling the information gap between the visual content of the images and their corresponding descriptions is a long-standing problem in image captioning. 
Based on the encoder-decoder pipeline \cite{vinyals2015show,yang2016review,you2016image}, much progress has been made on image captioning. 
For example, \cite{xu2015show} introduces the visual attention that adaptively attends to the salient areas in the image, 
\cite{lu2017knowing} proposes an adaptive attention model that decides whether to attend to the image or to the visual sentinel, 
\cite{yang2019auto} corporates learned language bias as a language prior for more human-like captions, 
\cite{luo2018discriminability} and \cite{guo2019mscap} focus on the discriminability and style properties of image captions respectively, 
and \cite{Rennie2016Self} adopts reinforcement learning (RL) that directly optimize evaluation metric. 

Recently, some works have been proposed to encode more discriminative visual information into captioning models. 
For instances, Up-Down \cite{anderson2017bottom} extracts region-level image features for training, 
\cite{Yao2016Boosting} incorporates image-level attributes into the encoder-decoder framework by 
training a Multiple Instance Learning \cite{Fang2014From} based attribute detectors. 
However, all these works focus on representing visual content with either objects/grids or global attributes, 
but fail to model object-object relationships. 
Differently, our method simultaneously models objects, the instance-level attributes, and relationships with structured graph of VSUs.

\begin{figure*}[!ht] 
	\centering
	\includegraphics[width=7in]{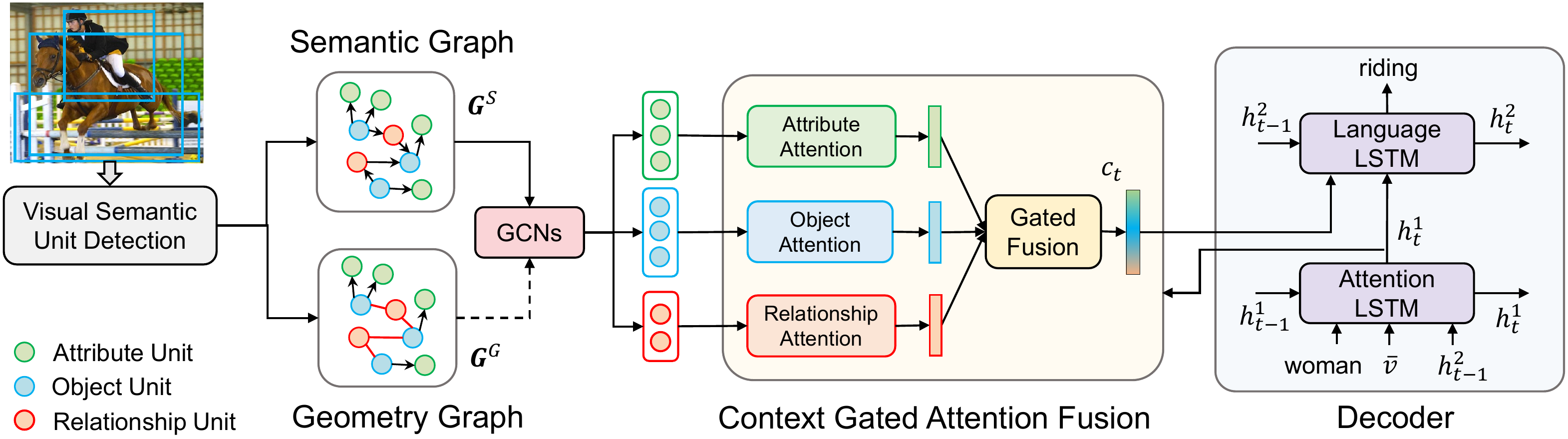}
	\caption{
		Overview of our method. Given an image, we represent it as structured graphs of visual semantic units (objects, attributes, and relationships in the image) and generate a caption based on them. 
	}
	\label{fig:framework}
\end{figure*}

\paragraph{Scene Graphs Generation and GCNs. } 
Recently, inspired by representations studied by the graphics community, 
\cite{johnson2015image} introduced the problem of generating scene graphs from images, 
which requires detecting objects, attributes, and relationships of objects. 
Many approaches have been proposed for the detection of both objects and their relationships \cite{xu2017scene,yang2018graph,zellers2018neural}. 
Recently, some works have been proposed that leverage scene graph for improving scene understanding in various tasks, \eg visual question answering \cite{teney2017graph}, referring expression understanding \cite{nagaraja2016modeling}, image retrieval \cite{johnson2015image}, and visual reasoning \cite{chen2018iterative}. 
In these works, GCNs \cite{kipf2016semi,gilmer2017neural,bastings2017graph} are often adopted to learn node embeddings. 
A GCN is a multilayer neural network that operates directly on a graph, in which information flows along edges of the graph. 

More similar to our work, GCN-LSTM \cite{yao2018exploring} refines the region-level features by leveraging object relationships and GCN. 
However, GCN-LSTM treats relationships as edges in the graph, which are implicitly encoded in the model parameters. 
While instead, our method considers relationships as additional nodes in the graph and thus can explicitly model relationships by 
learning instance-specific representations for them.

\section{Approach}

\subsection{Problem Formulation}
Image captioning models typically follow the encoder-decoder framework. 
Given an image $I$, the image encoder $\mathcal{E}$ is used to obtain the visual representation $V=\left\{v_{1}, \dots, v_{k}\right\}, v_{i} \in \mathbb{R}^{D}$, 
where each $v_i$ represents some features about the image content. 
Based on $V$, the caption decoder $\mathcal{D}$ generates a sentence $y$ by: 
\begin{equation}
V = \mathcal{E}(I), \ \ y = \mathcal{D}(V).
\end{equation}
The objective of image captioning is to minimize a cross entropy loss.

How to define and represent $V$ is a fundamental problem in this framework. 
In this work, we consider the visual content as structured combinations of the following three kinds of VSUs: 
\begin{itemize}
	\item Object units ($O$): the individual object instances in the image.
	\item Attribute units ($A$): the properties following each object.
	\item Relationship units ($R$): the interactions between object pairs. 
\end{itemize}
That is, we define $V=V_o\cup V_a\cup V_r$, where $V_o, V_a$, and $V_r$ denote the collections of visual representations for $O, A$, and $R$, respectively. 

The overall framework of our method is shown in Figure~\ref{fig:framework}. 
First, we detect $O$, $A$, and $R$ from the image with object, attribute, and relationship detectors respectively, 
based on which, a semantic graph and a geometry graph are constructed by regarding each VSU as the nodes and the connections between two VSUs as the edges.   
Then, GCNs are applied to learn context-aware embeddings for each of the nodes/VSUs. 
Afterward, a context gated attention fusion module is introduced to hierarchically align each word with the embedded VSUs, 
and the resulting context vector is fed into a language decoder for predicting words.

\subsection{Graph Representations of Visual Semantic Units}
\label{grah_build}

\paragraph{Visual Semantic Units Detection. }
We first detect the three types of VSUs by an object detector, an attribute classifier, and a relationship detector respectively, following \cite{yang2019auto}.  
Specifically, Faster R-CNN \cite{ren2015faster} is adopted as the object detector. 
Then, we train an attribute classifier to predict the instance attributes for each detected object, which is a simple multi-layer perceptron (MLP) network followed by a softmax function.   
MOTIFNET \cite{zellers2018neural} is adopted as the semantic relationship detector to detect pairwise relationships between object instances using the publicly available code\footnote{https://github.com/rowanz/neural-motifs}. 
Finally, we obtain a set of objects, attributes, and relationships, \ie the VSUs ($O$, $A$, and $R$). 
We denote $o_i$ as the $i$-th object, $a_{i,k}$ as the $k$-th attribute of $o_i$, 
and $r_{ij}$ as the relationship between $o_i$ and $o_j$. 
$<$$o_i, r_{ij}, o_j$$>$ forms a triplet, which means $<$subject, predicate, object$>$.

\begin{figure}[!t] 
	\centering
	\includegraphics[width=3in]{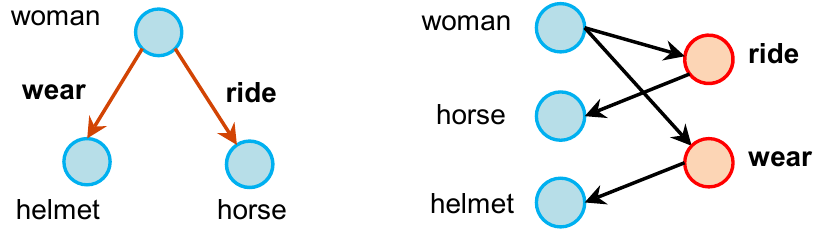}
	\vspace{-0.1cm}
	\caption{
		Comparison between regarding relationships as edges and as nodes in the graph. 
		The attribute nodes are omitted for clarity.   
	}
	\vspace{-0.4cm}
	\label{fig:edge}
\end{figure}

\paragraph{Graph Construction. }
It is intuitive to introduce graph as a structured representation of the VSUs, which contains nodes and edges connecting them.
We can naturally regard both objects and attributes as nodes in the graph, where each attribute node $a_{i,k}$ is connected with the object node $o_i$ with 
a directed edge ($o_i  \to a_{i,k}$), meaning ``object $o_i$ owns attribute $a_{i,k}$". 
However, whether to represent relationships as edges or nodes in the graph remains uncertain. 
A common and straightforward solution is to represent relationships as edges connecting pairwise object nodes in the graph \cite{yao2018exploring,teney2017graph} (the left side of Figure~\ref{fig:edge}.). 
However, such models only learn vector representation for each node (objects) and implicitly encode edge (relationships) information in the form of GCN parameters, while the relationship representations are not directly modeled. 

Ideally, relationships should have instance-specific representations, in the same way as objects, and they should also inform decisions made in the decoder. 
Thus, we propose to explicitly modeling relationship representations by turning the relationships as additional nodes in the graph. 
Concretely, for each relationship $r_{ij}$, we add a node in the graph and draw two directed edges: $o_i \to r_{ij}$ and $ r_{ij} \to o_j$. 
The subgraph of all $o_i$ and $r_{ij}$ is now a \textbf{bipartite}. 
An instance is shown in the right side of Figure \ref{fig:edge}. 

Formally, given three sets of object nodes (units) $O=\{o_i\}$, attribute nodes (units) $A=\{a_{i,k}\}$, 
and relationship nodes (units) $R=\{r_{ij}\}$, we define the graph $\mathcal{G} =(\mathcal{N}, \mathcal{E})$ as comprehensive representation of the visual content for the image, 
where $\mathcal{N} = O \cup A  \cup R$ is the nodes and $\mathcal{E}$ is a set of directed/undirected edges.

\paragraph{Semantic Graph and Geometry Graph. }
We consider two types of visual relationships between objects in the image: semantic relationship and geometry relationship, 
which result in two kinds of graphs, \ie \textbf{semantic graph $\mathcal{G}^S$ and geometry graph  $\mathcal{G}^G$}. 
The semantic relationship $R^S$ unfolds the inherent action/interaction between objects, \eg ``woman riding horse". 
Semantic relationships are detected using MOTIFNET. 
Geometry relationship $R^G$ between objects is an important factor in visual understanding that connects isolated regions together, \eg ``woman on horse". 
Here we consider the \textit{undirected} relative spatial relationships $<$$o_{i^\prime}, r_{i^\prime j^\prime}, o_{j^\prime}$$>$, where $o_{i^\prime}$ is the one with larger size between $o_{i}$ and $o_{j}$, while $o_{j^\prime}$ is the smaller one. 
Instead of using a fully connected graph, we assign geometry relationships between two objects 
if their Intersection over Union (IoU) and relative distance are within given thresholds.  

\subsection{Embedding Visual Semantic Units with GCN}
\label{grah_embed}

\paragraph{Node Features. } 
We integrate three kinds of content cues for $o_i, a_{i,k}$, and $r_{ij}$, \ie 
visual appearance cue, semantic embedding cue, and geometry cue. 

\textit{Visual Appearance Cue.}
Visual appearance contains delicate details of the object and is a key factor for image understanding. 
We use the region-level RoI feature from Faster R-CNN as the visual appearance cue for each object, which is a 2048-dimensional vector, denoted as $\boldsymbol{f}^{v}_{o_i}$. 

\textit{Semantic Embedding Cue.}
The content of objects, attributes and relationships can largely be described with their categories. 
Thus, we introduce their category labels as the semantic information, which are obtained from the outputs of object, attribute and relationship detectors. 
Specifically, we use three independent and trainable word embedding layers to maps the object categories into feature embedding vectors (denoted as $\boldsymbol{f}^s_{o_i}, \boldsymbol{f}^s_{a_{i,k}}$, and $\boldsymbol{f}^s_{r_{ij}}$) for $o_i, a_{i,k}$, and $r_{ij}$, respectively. 

\textit{Geometry Cue.}
Geometry information is complementary to visual appearance information, 
as it reflects the spatial patterns of individual objects or the relation between two objects. 
Denote the box of a localized object $o_i$ as $(x_i, y_i, w_i, h_i )$, where $(x_i, y_i)$ are the coordinates of the center of the box, 
and $(w_i, h_i)$ are its width and height, and denote the width and height of the image as $(w, h)$.
We encode the relative geometric cue of $o_i$ with a 5-dimensional vector:
\begin{equation}
\boldsymbol{f}^{g}_{o_i} =\big[\frac{x_i}{w}, \frac{y_i}{h}, \frac{w_i}{w},  \frac{h_i}{h}, \frac{w_i h_i}{w h} \big]. 
\end{equation}
We encode the geometric cue of each relationship $r_{ij}$ with a 8-dimensional vector:

\begin{align}
\boldsymbol{f}^{g}_{r_{ij}}  = &
\Big[
\underbrace{  \frac{x_j-x_i}{\sqrt{w_ih_i}}, \frac{y_j-y_i}{\sqrt{w_ih_i}} }_{r_1}, \  
\underbrace{  \frac{w_j}{w_i}, \frac{h_j}{h_i}, \frac{w_jh_j}{w_ih_i}      }_{r_2},  \  
\underbrace{  \frac{o_i\cap o_j}{o_i\cup o_j}         }_{r_3}, \   \nonumber \\ 
& \underbrace{  \frac{\sqrt{(x_j-x_i)^2 +(y_j-y_i)^2}}{\sqrt{w^2+h^2}}   }_{r_4}, \ 
\underbrace{  atan2(\frac{y_j-y_i}{x_j-x_i})          }_{r_5} 
\Big], 
\label{eqn:spatial_feat}
\end{align}
where $r_1$ represents the normalized translation between the two boxes,  
$r_2$ is the ratio of box sizes, 
$r_3$ is the IoU between boxes, 
$r_4$ is the relative distance normalized by diagonal length of the image, 
$r_5$ is the relative angle $\theta\in (-\pi, \pi] $ between the two boxes.

We fuse the visual appearance cues, semantic embedding cues, and geometry cues to 
obtain the features for each node/VSU.
We aggregate the top-$N_a$ predicted attributes $[ a_{i,1} ,  ..., a_{i,N_a} ]$ (sorted by the classification confidences) of $o_i$ into a single attribute unit $a_i$ for each object. 
Denote the features corresponding to $o_i, a_{i,k}$, and $r_{ij}$ nodes as 
$\boldsymbol{f}_{o_i}, \boldsymbol{f}_{a_{i}}$, and $\boldsymbol{f}_{r_{ij}}$, respectively, and use superscripts $\boldsymbol{f}^G$ and $\boldsymbol{f}^S$ to denote the features of $\mathcal{G}^{S}$ and $\mathcal{G}^{G}$. 
The feature of $o_i$, $a_i$, $r_{ij}$ are given by: 
\begin{align}
&\boldsymbol{f}_{o_i} = \phi_o ([ \boldsymbol{f}^{v}_{o_i}; \boldsymbol{f}^{s}_{o_i} ; \boldsymbol{f}^{g}_{o_i} ]), \\ \label{eqn:cueo}
&\boldsymbol{f}_{a_i} = \phi_a ( [\boldsymbol{f}^s_{a_{i,1}}; \ldots ; \boldsymbol{f}^s_{a_{i,N_a}}]), \\
&\boldsymbol{f}^S_{r_{ij}} = \phi_r (\boldsymbol{f}^{s}_{r_{ij}}),\\
&\boldsymbol{f}^G_{r_{ij}} = \phi_r (\boldsymbol{f}^{g}_{r_{ij}}), 
\end{align}
where $[*;*]$ means concatenation operation, and $\phi_o , \phi_a$, and $\phi_r $ are feature projection layers which are all implemented as FC-ReLU-Dropout. 

\paragraph{Node Embedding with GCNs. }
After obtaining the features for the three kinds of nodes/VSUs, 
we then adopt GCNs: $g_{o}, g_{a}$, and $g_{r}$, 
to learn semantic (geometrical) context-aware embeddings for the $o_{i}, a_i$, and $r_{i j}$ nodes in $\mathcal{G}$, respectively.

The object unit embedding $\boldsymbol{u}_{o_i}$ is calculated as: 
\begin{equation}
\boldsymbol{u}_{o_i}= g_o(\boldsymbol{f}_{o_{i}}) + \boldsymbol{f}^{v}_{o_i}, 
\label{eqn:feato}
\end{equation}
where adding $\boldsymbol{f}^{v}_{o_i}$ serves as a kind of ``residual connection", 
which we empirically found helpful for model performance. 

Given the feature $\boldsymbol{f}_{a_i}$ of an attribute unit $a_i$, 
we integrate it with its object context $o_i$ for calculating the attribute unit embedding $\boldsymbol{u}_{a_i}$ by:
\begin{equation}
\boldsymbol{u}_{a_{i}}= g_{a}\left(\boldsymbol{u}_{o_{i}}, \boldsymbol{f}_{a_{i}}\right) +  \boldsymbol{f}_{a_{i}}.
\label{eqn:feata}
\end{equation}

For each relationship $r_{ij}$ and its associated relationship triplet $<$$o_i, r_{ij}, o_j$$>$, 
we aggregate the information from its neighboring nodes ($o_i$ and $o_j$) to calculate the relationship unit embedding $u_{r_{ij}}$ by:
\begin{equation}
\boldsymbol{u}_{r_{i j}}=g_{r}\left(\boldsymbol{u}_{o_{i}}, \boldsymbol{f}_{r_{i j}}, \boldsymbol{u}_{o_{j}}\right) + \boldsymbol{f}_{r_{i j}}. 
\label{eqn:featr}
\end{equation}

In our implementation, $g_{o}, g_{a}$, and $g_{r}$ use the same structure with independent parameters: a concatenation operation followed by a FC-ReLU-Dropout layer. Note that for $\mathcal{G}^S$ and $\mathcal{G}^G$, their $\boldsymbol{u}_{o_{i}}$ and $\boldsymbol{u}_{a_{i}}$ are the same, while 
their $\boldsymbol{u}_{r_{i j}}$ are independently calculated and denoted as $\boldsymbol{u}^S_{r_{i j}}$ and $\boldsymbol{u}^G_{r_{i j}}$ respectively. 
We will introduce in Sec. \ref{align} about how to fuse $\mathcal{G}^S$ and $\mathcal{G}^G$ by leveraging all $\boldsymbol{u}_{o_{i}}, \boldsymbol{u}_{a_{i}}, \boldsymbol{u}^S_{r_{i j}}$, and $\boldsymbol{u}^G_{r_{i j}}$.

\subsection{Aligning Textual Words with Visual Semantic Units}
\label{align}
We next discuss how to effectively integrate the learned embeddings of various types of VSUs 
into sequence generation network for image captioning. 
\paragraph{Context Gated Attention for Word-Unit Alignment. }
We have two observations about the correlation between linguistic words and VSUs: 
1) both a word in the caption and a VSU can be assigned into one of the three categories: objects, attributes, and relationships, 
2) a word often could be aligned with one of the VSUs in the image, which convey the same information in different modalities. 
Starting from the two observations, we design a context gated attention module (CGA) to hierarchically align each caption word with the VSUs by soft-attention mechanism. 

Specifically, at each step, CGA first performs three independent soft-attentions for VSUs in the three categories, \ie object, attribute, and relationship, respectively. 
Mathematically, given the attention query $\boldsymbol{h}^1_t$ from the decoder at the $t$-th step, we calculate: 
\begin{align}
att^O_t &= ATT^O(\boldsymbol{h}^1_t, \{\boldsymbol{u}_{o_i}\}), \\
att^A_t &= ATT^A(\boldsymbol{h}^1_t, \{\boldsymbol{u}_{a_i}\}), \\
att^R_t &= ATT^R(\boldsymbol{h}^1_t, \{\boldsymbol{u}_{r_{ij}}\}), 
\label{eqn:att}
\end{align}
where $ATT^O$, $ATT^A$, and $ATT^R$ are soft-attention functions with independent parameters, 
while $att^O_t$, $att^A_t$, and $att^R_t$ are the resulting context vectors for object, attribute, and relationship units, respectively. 
We implement $ATT^O$, $ATT^A$, and $ATT^R$ with the same structure as proposed in \cite{xu2015show}. 
The attention function $att_{t} = ATT (\boldsymbol{h}^1_t, \{\boldsymbol{u}_{k}\})$ is calculated by:
\begin{align}
a_{k, t} &=\boldsymbol{w}_{a}^{T} \tanh \left(\boldsymbol{W}_{a} [ \boldsymbol{u}_{k}; \boldsymbol{h}^1_{t} ] \right), \\ 
\boldsymbol{\alpha}_{t} &=\operatorname{softmax}\left(\boldsymbol{a}_{t}\right), \\
att_{t} &=\sum_{k=1}^{K} \alpha_{k, t} \boldsymbol{u}_{k},
\end{align}
where $\alpha_{k, t}$ is a normalized attention weight for each of the unit embedding vectors $\boldsymbol{u}_k$, 
$att_{t}$ is the aggregated result, and $W_{a} \text { and } w_{a} $ are learnable parameters.

Given $att^O_t$, $att^A_t$, and $att^R_t$, a \textit{gated fusion} process is then performed at the higher category level of VSUs. 
Concretely, we generate a probability for each of the three VSU categories: object, attribute, and relationship as follows:
\begin{align}
\boldsymbol{\beta}_{t} &=sigmoid \left(\boldsymbol{W}_{\beta} [ \boldsymbol{h}^1_{t}; att^O_t; att^A_t; att^R_t ] \right), 
\label{eqn:gate}
\end{align}
where $\boldsymbol{\beta}_{t}\in \mathbb{R}^{3 \times 1}$ is the gating weights for each category, $W_{\beta} $ are learnable parameters. 
Denote the gating weights for object, attribute, and relationship categories as $\beta^o_t, \beta^a_t , \beta^r_t $, respectively. 
Then $\boldsymbol{\beta}_{t} = [\beta^o_t, \beta^a_t , \beta^r_t]$ indicates which VSU category the current word is about (object, attribute, or relationships),  
and decides which VSU category should be paid more attention currently. 
Then we compute the current context vector by aggregating $att^O_t$, $att^A_t$, and $att^R_t$ according to $\boldsymbol{\beta}_{t}$:
\begin{align}
\boldsymbol{c}_{t} &=concat(\beta^o_t  att^O_t  ,  \beta^a_t att^A_t,  \beta^r_t att^R_t ).
\label{eqn:contextvec}
\end{align}
In order to utilize the unit embeddings learned from both $\mathcal{G}^S$ and $\mathcal{G}^G$, 
we extend the calculation of $\boldsymbol{c}_{t}$ to include the attentional results of both $\mathcal{G}^S$ and $\mathcal{G}^G$. 
Specifically, denote the $att^R_t, \beta^r_t$ of $\mathcal{G}^S$ and $\mathcal{G}^G$ as $att^{RS}_t, att^{RG}_t$ and $\beta^{rs}_t, \beta^{rg}_t$, respectively. 
Then we calculate $\boldsymbol{c}_{t}$ as:
\begin{align}
\boldsymbol{c}_{t} &=concat(\beta^o_t att^O_t  ,  \beta^a_t att^A_t,  \beta^{rs}_t att^{RS}_t, \beta^{rg}_t att^{RG}_t ).
\label{eqn:g+g}
\end{align}

The context vector $\boldsymbol{c}_{t}$ is then ready to be used by the sentence decoder for predicting the $t$-th word.

\paragraph{Attention based Sentence Decoder. }
We now introduce our sentence decoder which is a two-layer Long Short-Term Memory (LSTM) \cite{hochreiter1997long} network with our context gated attention module injected in the middle it, as is shown in the right part of Figure \ref{fig:framework}. 
Following \cite{anderson2017bottom}, the input vector to the first LSTM (called attention LSTM)
at each time step is the concatenation of the embedding of the current word, the the mean-pooled image feature $\overline{\boldsymbol{v}}=\frac{1}{k} \sum_{i} \boldsymbol{f}_{o_i}^v$, as well as the previous hidden state of the second LSTM, $\boldsymbol{h}^2_{t-1}$. 
Hence the updating procedure for the first LSTM is given by:
\begin{equation}
\boldsymbol{h}_t^1 = \text{LSTM}^1\left(\boldsymbol{h}^1_{t-1};  \left[\boldsymbol{h}_{t-1}^{2}; \overline{\boldsymbol{v}}; \boldsymbol{W}_{e} \boldsymbol{\Pi}_{t}\right] \right),
\end{equation}
where $\boldsymbol{W}_{e} \in \mathbb{R}^{E \times N}$ is a word embedding matrix, and $\boldsymbol{\Pi}_{t}$ is one-hot encoding vector of the current input word. 
$\boldsymbol{h}_t^1$ is leveraged as the query for the context gated attention module (Eqn.\ \ref{eqn:att}) to obtain the current context vector $\boldsymbol{c}_{t}$. 

The second LSTM (called language LSTM) takes as input the context vector $\boldsymbol{c}_{t}$ and the
hidden state of the first LSTM. Its updating procedure is given by:
\begin{equation}
\boldsymbol{h}_t^2 = \text{LSTM}^2\left(\boldsymbol{h}^2_{t-1},  \left[\boldsymbol{h}_{t}^{1}; \boldsymbol{c}_t \right] \right). 
\end{equation}
$\boldsymbol{h}_t^2$ is then used to predict the next word $y_{t}$ through a softmax layer. 

\section{Experiments}
\subsection{Datasets and Evaluation Metrics} 
\noindent\textbf{MS-COCO  }\cite{lin2014microsoft}. The dataset is the most popular benchmark for image captioning. 
We use the `Karpathy' splits [19] that have been used extensively for
reporting results in prior works. This split contains 113,287
training images with five captions each, and 5k images for validation and test splits, respectively. 
We follow standard practice and perform only minimal
text pre-processing: converting all sentences to lower case,
tokenizing on white space, discarding rare words which occur less than 5 times, and trimming each caption to a maximum of 16 words,
resulting in a final vocabulary of 9,487 words.

\noindent\textbf{Evaluation Metrics. }
We use the standard automatic evaluation metrics --- BLEU-1,2,3,4, METEOR, ROUGE-L, CIDEr \cite{lin2014microsoft}, and SPICE \cite{Anderson2016SPICE} ---  to evaluate caption quality, 
denoted as B@N, M, R, C and S, respectively.

\label{sec:detail}
\subsection{Implementation Details} 

Visual Genome (VG) \cite{krishna2017visual} dataset is exploited to train our object detector, attribute classifier, and relationship detector. 
We use three vocabularies of 305 objects, 103 attributes, and 64 relationships respectively, following \cite{yang2019auto}. 
For the object detector, we use the pre-trained Faster R-CNN model along with ResNet-101 \cite{he2016deep} backbone provided by \cite{anderson2017bottom}. 
We use the top-3 predicted attributes for attribute features, \ie $N_a=3$. For constructing the geometry graph, we consider two objects to have interactions 
if their boxes satisfy two conditions: $r_2 < 0.2$ and $r_4<0.5$, 
where $r_2$ and $r_4$ are the IoU and relative distance in Eqn.\ \ref{eqn:spatial_feat}. 

The number of output units in the feature projection layers ($\phi_o, \phi_a, \phi_r $) and 
the GCNs ($g_o, g_a, g_r $) are all set to 1000. 
For the decoder, we set the number of hidden units in each LSTM and each attention layer to 1,000 and 512, respectively. 
We use four independent word embedding layers for embedding input words, objects, attributes, and relationships categories, 
with the word embedding sizes set to 1000, 128, 128, and 128, respectively. 
We first train our model under a cross-entropy loss using Adam \cite{kingma2014adam} optimizer and the learning rate was initialized to
$5\times 10^{-4}$ and was decayed by $0.8$ every 5 epochs.
After that, we train the model using reinforcement learning \cite{Rennie2016Self} by optimizing the CIDEr reward.
When testing, beam search with a beam size of 3 is used.

\subsection{Ablation Studies}
We conduct extensive ablative experiments to compare our model against alternative architectures and settings, as shown in Table \ref{ablation}. 
We use \textit{Base} to denote our baseline model,  
which is our implementation of \textit{Up-Down} \cite{anderson2017bottom}. 

\begin{table}[tbp]
	\centering
	\caption{Ablations of our method, evaluated on MS-COCO Karpathy split.}
	\vspace{-0.2cm}
	\label{ablation}
	\begin{tabular}{clrrrrr}
		\toprule
		& Model & \multicolumn{1}{c}{B@4\newline{}} & \multicolumn{1}{c}{M\newline{}} & \multicolumn{1}{c}{R} & \multicolumn{1}{c}{C} & \multicolumn{1}{c}{S\newline{}} \\
		\midrule
		\multicolumn{1}{l}{ } & Base   & 36.7  & 27.9  & 57.5  & 122.8  & 20.9  \\ 
		\midrule
		\multirow{11}[2]{*}{(a)} 
		& $O$ &  36.9  & 27.9  & 57.6  & 123.8  & 21.1  \\
		& $A$ &   36.9  & 27.8  & 57.6  & 123.4  & 21.1  \\
		& $R^S$ &  35.7  & 27.5  & 57.1  & 119.3  & 21.2  \\
		& $R^G$ &  36.6  & 27.7  & 57.4  & 121.5  & 21.1  \\
		& $OA$ &  37.6  & 28.1  & 58.0  & 126.1  & 21.7  \\
		& $OR^S$ &  37.7  & 28.2  & 58.1  & 126.3  & 21.9  \\
		& $OR^G$ &  37.8  & 28.2  & 57.9  & 126.5  & 21.9  \\
		& $AR^S$ &  37.3  & 28.1  & 57.7  & 125.9  & 21.7  \\
		& $OAR^S$($\mathcal{G}^S$) & 37.9  & 28.3  & 58.2  & 127.2  & 21.9  \\
		& $OAR^G$($\mathcal{G}^G$) &  38.0  & 28.3  & 58.1  & 127.2  & 21.9  \\
		& $\mathcal{G}^S+ \mathcal{G}^G $ &  \bftab{38.4}  & \bftab{28.5}  & \bftab{58.4}  & \bftab{128.6}  & \bftab{22.0}  \\
		\midrule
		\multirow{2}[2]{*}{(b)} 
		& $\mathcal{G}^S$+gate &   38.1  & 28.4  & 58.1  & 128.1  & 22.0  \\
		& $\mathcal{G}^G$+gate &   38.3  & 28.4  & 58.3  & 128.2  & 22.0  \\
		\midrule
		\multirow{4}[2]{*}{(c)} 
		& baseline: $\mathcal{G}^G$+gate   &   \bftab{38.3}  & \bftab{28.4}  & 58.3  & \bftab{128.2}  & \bftab{22.0}  \\
		&   w/o $f^v_{o_i}$ &   32.8  & 26.3  & 55.4  & 111.9  & 19.7  \\
		&   w/o $f^s_{o_i}$ &   37.9  & 28.3  & 58.2  & 126.9  & 21.8  \\
		&   w/o $f^g_{o_i}$ &   \bftab{38.3}  & \bftab{28.4}  & \bftab{58.4}  & 127.7  & \bftab{22.0}  \\
		\midrule
		\multirow{2}[2]{*}{(d)} 
		& $\mathcal{G}^G$+shareAtt. &   37.4  & 28.0  & 57.9  & 124.1  & 21.3  \\
		& Base+multiAtt. &  37.1  & 27.9  & 57.7  & 123.7  & 21.3  \\
		\bottomrule
	\end{tabular}%
	\vspace{-0.1cm}
	\label{tab:ablation}%
\end{table}%

\begin{table*}[tbp]
	\centering
	\caption{Leaderboard of the published state-of-the-art image captioning models on the online COCO test server, where c5 and c40 denote using 5 and 40 references for testing, respectively. 
	}
	\vspace{-0.2cm}
	\begin{tabular}{@{\extracolsep{3pt}}@{\kern\tabcolsep}lrrrrrrrrrrrrrr}
		\toprule
		Model & \multicolumn{2}{c}{BLEU-1} & \multicolumn{2}{c}{BLEU-2} & \multicolumn{2}{c}{BLEU-3} & \multicolumn{2}{c}{BLEU-4} & \multicolumn{2}{c}{METEOR} & \multicolumn{2}{c}{ROUGE-L} & \multicolumn{2}{c}{CIDEr} \\
		\cmidrule{2-3}  \cmidrule{4-5} \cmidrule{6-7} \cmidrule{8-9} \cmidrule{10-11} \cmidrule{12-13} \cmidrule{14-15} %
		Metric & \multicolumn{1}{c}{c5} & \multicolumn{1}{c}{c40} & \multicolumn{1}{c}{c5} & \multicolumn{1}{c}{c40} & \multicolumn{1}{c}{c5} & \multicolumn{1}{c}{c40} & \multicolumn{1}{c}{c5} & \multicolumn{1}{c}{c40} & \multicolumn{1}{c}{c5} & \multicolumn{1}{c}{c40} & \multicolumn{1}{c}{c5} & \multicolumn{1}{c}{c40} & \multicolumn{1}{c}{c5} & \multicolumn{1}{c}{c40} \\
		\midrule
		SCST \cite{Rennie2016Self} & 78.1  & 93.7  & 61.9  & 86.0  & 47.0  & 75.9  & 35.2  & 64.5  & 27.0  & 35.5  & 56.3  & 10.7  & 114.7  & 116.0  \\
		LSTM-A \cite{Yao2016Boosting} & 78.7  & 93.7  & 62.7  & 86.7  & 47.6  & 76.5  & 35.6  & 65.2  & 27.0  & 35.4  & 56.4  & 70.5  & 116.0  & 118.0  \\
		StackCap \cite{gu2017stack} & 77.8  & 93.2  & 61.6  & 86.1  & 46.8  & 76.0  & 34.9  & 64.6  & 27.0  & 35.6  & 56.2  & 70.6  & 114.8  & 118.3  \\
		Up-Down \cite{anderson2017bottom} & \bftab{80.2}  & \bftab{95.2}  & 64.1  & \bftab{88.8}  & 49.1  & 79.4  & 36.9  & 68.5  & 27.6  & 36.7  & 57.1  & 72.4  & 117.9  & 120.5  \\
		CAVP \cite{liu2018context}  & 80.1  & 94.9  & \bftab{64.7}  & \bftab{88.8}  & \bftab{50.0}  & \bftab{79.7}  & \bftab{37.9}  & \bftab{69.0}  & 28.1  & 37.0  & \bftab{58.2}  & \bftab{73.1}  & 121.6  & 123.8  \\
		Ours: VSUA & 79.9  & 94.7  & 64.3  & 88.6  & 49.5  & 79.3  & 37.4  & 68.3  & \bftab{28.2}  & \bftab{37.1}  & 57.9  & 72.8  & \bftab{123.1}  & \bftab{125.5}  \\
		\bottomrule
	\end{tabular}%
	\label{tab:server}%
\end{table*}%

\begin{table}[tp]
	\centering
	\caption{Performance comparison with the existing methods on MS-COCO Karpathy test split. }
	\vspace{-0.2cm}
	\begin{tabular}{lrrrrrr}
		\toprule
		Approach\newline{}  & \multicolumn{1}{c}{B@4\newline{}} & \multicolumn{1}{c}{M\newline{}} & \multicolumn{1}{c}{R} & \multicolumn{1}{c}{C} & \multicolumn{1}{c}{S\newline{}} \\
		\midrule
		Google NICv2 \cite{vinyals2017show}   & 32.1  &   25.7 & -- & 99.8  & -- \\
		Soft-Attention \cite{xu2015show}  & 25.0  & 23.0  & --     & --     & -- \\
		LSTM-A \cite{Yao2016Boosting} & 32.5  & 25.1  & 53.8  & 98.6  & --  \\
		Adaptive \cite{lu2017knowing}  & 33.2  & 26.6  & --     & 108.5  & 19.4 \\
		\midrule
		SCST \cite{Rennie2016Self}   & 33.3  & 26.3  & 55.3  & 111.4  & -- \\
		Up-Down  \cite{anderson2017bottom} & 36.3  & 27.7  & 56.9  & 120.1  & 21.4  \\
		Stack-Cap \cite{gu2017stack}  & 36.1  & 27.4  & 56.9  & 120.4  & 20.9  \\
		CAVP \cite{liu2018context} & \bftab{38.6} & 28.3 & \bftab{58.5} & 126.3 & 21.6 \\
		GCN-LSTM \cite{yao2018exploring}  & 38.2  & \bftab{28.5}  & 58.3  & 127.6  & \bftab{22.0}  \\
		\midrule
		Ours: VSUA  & 38.4  & \bftab{28.5}  & 58.4  & \bftab{128.6}  & \bftab{22.0}  \\
		\bottomrule
	\end{tabular}%
	\vspace{-0.3cm}
	\label{tab:test}%
\end{table}%

\paragraph{(a) How much does each kind of VSUs contribute? }
The experiments in Table \ref{ablation}(a) answer this question, 
where we compare the model performance of using different combinations of the three categories of VSUs. 
We first denote the object, attribute, and relationship units as $O$, $A$, $R^S$ ($R^G$), respectively. 
$R^S$ means the semantic relationship and $R^G$ in $\mathcal{G}^S$ is the geometry relationship in $\mathcal{G}^G$. 
Then we denote the various combinations of the input components ($att^O_t$, $att^A_t$, and $att^R_t$) to be used in Eqn.\ \ref{eqn:contextvec} as the  combinations of the corresponding symbols: $O$, $A$, $R^S$ ($R^G$). 
We have the following observations from Table \ref{ablation} (a) . 

\textbf{First}, we look at the $O$ model,  
where we simply replace the original visual features $\boldsymbol{f}^v_{o_i}$ in the baseline model with our embeddings of object units ($\boldsymbol{u}_{o_i}$ in Eqn.\ \ref{eqn:feato}). We can see that this simple modification brings slight improvement on model performance, indicating the effectiveness of fusing the visual appearance cues, semantic embedding cues, and geometry cues for representing object units. 
\textbf{Second}, results of the $A$, $R^S$, $R^G$ models show that using attribute or relationship units alone result in decreased model performance. 
That is because although the computation of their embeddings have involved the embeddings of the object units (\eg Eqn. \ref{eqn:feata}), 
the added residual connections make $\boldsymbol{f}_{a_i}$ ($\boldsymbol{f}_{r_i}$) the dominant factors in the computation process. 
It is also noteworthy that all of $A$, $R^S$, $R^G$ achieve comparable results as to the baseline, indicating the effectiveness of their learned unit embeddings. 
\textbf{Third}, the results of $OA$, $OR^S$, $OR^G$ models represent significant outperform that of the $O$ model.  
The performance of $AR^S$ is better than $R^S$, however is inferior to $OR^S$, 
which again shows the importance of object units. 
\textbf{Fourth}, Combining object, attribute, and relationship units altogether, \ie $OAR^S$ and $OAR^G$, brings the highest performance. 
The results show that the three kinds of VSU are highly complementary to each other.  
\textbf{Finally}, we combine VSUs from both the spatial graph $\mathcal{G}^S$ and the geometry graph $\mathcal{G}^G$ for training, denoted as $\mathcal{G}^S+\mathcal{G}^G$, which is also equivalent to $OAR^SR^G$.  
We see that $\mathcal{G}^S+\mathcal{G}^G$ further improves the performance over $\mathcal{G}^S$ and $\mathcal{G}^G$, indicating the compatibility between $R^S$ and $R^G$.

\paragraph{(b) The effect of context gated attention. } 
In the above experiments, the gating weights ($\boldsymbol{\beta}_t$) are set to 1. 
We now further apply our gated fusion process (Eqn.\ \ref{eqn:gate}) upon them, whose results are shown in Table \ref{ablation}(b). 
We can see that the performance of both $OAR^S$ and $OAR^G$ is further improved,  
showing that the hierarchical alignment process is beneficial to take full advantage of the VSUs. 
Overall, compared to the baseline, the CIDEr score is boosted up from $122.8$ to $128.6$.

\paragraph{(c) The effects of different content cues. } 
Using $\mathcal{G}^G+gate$ as the baseline, we discard (denoted as \textit{w/o}) each of the visual appearance cue $f^v_{o_i}$, semantic embedding cue $f^s_{o_i}$, and geometry cue $f^g_{o_i}$ 
from the computation process of $f_{o_i}$ (Eqn.\ \ref{eqn:cueo}). 
Note that, in \textit{w/o $f^v_{o_i}$}, we remove \textit{all} visual appearance features (including $f^v_{oi}$ and $\overline{v}$) from the sentence decoder. 
We can see that removing any of them results in decreased performance.  
Particularly, \textit{w/o $f^v_{o_i}$} only achieves a CIDEr score of $111.9$, 
indicating visual appearance cues are still necessary for image captioning.

\paragraph{(d) Does the improvement come from more parameters or computation? } 
First, in the \textit{$\mathcal{G}^G$+shareAtt.} model, instead of using three independent attention modules for $O$, $A$, $R$ in the $\mathcal{G}^G$ model, we use a \textit{single} attention module for aggregating their embeddings. We see that \textit{$\mathcal{G}^G$+shareAtt.} deteriorates the performance. 
Second, in the \textit{Base+multiAtt.} model, we replace the attention module in the Base model with three attention modules, which have the same structure and inputs but independent parameters. We can see that the performance of \textit{Base+multiAtt.} is far worse than our $\mathcal{G}^S$ and $\mathcal{G}^G$ models, 
although their decoders have similar numbers of parameters. 
The comparisons indicate that effect of our method is beyond increasing computation and network capacity.

\begin{figure}[!tp] 
	\centering
	\includegraphics[width=3in]{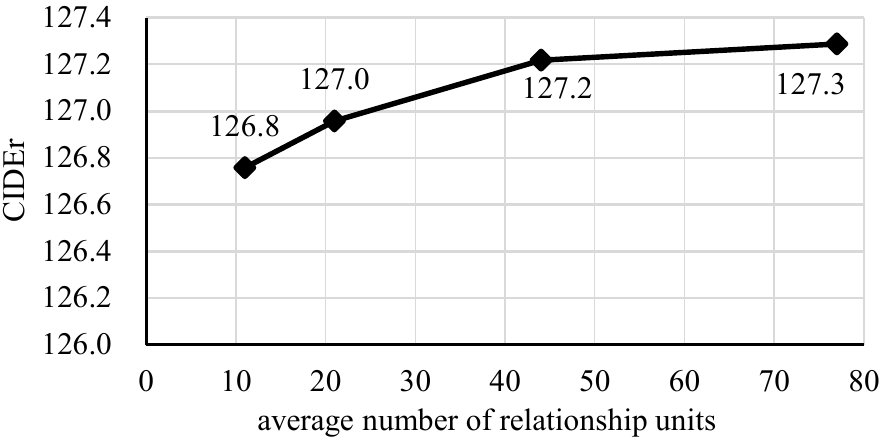}
	\vspace{-0.3cm}
	\caption{ Results of training our model with different numbers of relationship units for images. 	  		
	}
	\vspace{-0.4cm}
	\label{fig:num_box}
\end{figure}

\paragraph{(e) How many relationship units to use? }
We compare the effect of using various numbers of geometry relationship units for training. 
We have introduced in Section \ref{sec:detail} that we consider two objects to have interactions if $r_2 < 0.2$ and $r_4<0.5$, where $r_2$ means IoU.  
Thus, we adjust the threshold value $\gamma$ for $r_2$ to change the number of geometry relationship units for each image. 
Specifically, we set $\gamma$ to $0.1, 0.2, 0.3, 0.4$ respectively, 
which result in an average of $77.3, 43.8, 20.8, 11.1$ relationship units per image respectively. 
We then separately train the $\mathcal{G}^G+gate$ model for each of them. 
The changes of the CIDEr score are shown in Figure \ref{fig:num_box}. 
As we can see, basically, as the number of relationship units increases, 
the CIDEr score gradually increases. However, the performance differences are not very significant. 
Consider the trade-off between computation and model performance, 
we set $\gamma$ to $0.2$.

\subsection{Comparisons with State-of-The-Arts}
We compare our methods with Google NICv2 \cite{vinyals2017show}, Soft-Attention \cite{xu2015show}, LSTM-A \cite{Yao2016Boosting}, Adaptive \cite{lu2017knowing}, SCST \cite{Rennie2016Self}, StackCap \cite{gu2017stack}, Up-Down \cite{anderson2017bottom}, CAVP \cite{liu2018context}, and GCN-LSTM \cite{yao2018exploring}. 
Among them, LSTM-A incorporates attributes of the whole image for captioning,  
SCST uses reinforcement learning for training, Up-Down is the baseline with the same decoder as ours, 
Stack-Cap adopts a three-layer LSTM and more complex reward, CAVP models the visual context over time, 
and GCN-LSTM treats visual relationships as edges in a graph to help refining the region-level features.  
We name our $\mathcal{G}^S+ \mathcal{G}^G $ model as \textbf{VSUA} (\textit{Visual Semantic Units Alignment}) for convenience.

We show in Table~\ref{tab:test} the comparison between our \textit{single} model and state-of-the-art single-model methods on the MS-COCO Karpathy test split. 
We can see that our model achieves a new state-of-the-art score on CIDEr ($\bf{128.6}$), and comparable scores with GCN-LSTM and CAVP on the other metrics. 
Particular, relative to the Up-Down baseline, we push the CIDEr from $120.1$ to $128.6$.  
It is noteworthy that GCN-LSTM uses a considerable large batch size of 1024 and training epochs of 250, 
which are far beyond our computing resource and also larger than that of ours and the other methods (typically both are within 100). 
Table~\ref{tab:server} reports the performances of our single model \textit{without} any ensemble on the official MS-COCO evaluation server (by the date of 08/04/2019).
We can see that our approach achieves very competitive performance,
compared to the state-of-the-art.

\subsection{Qualitative Analysis}

\begin{figure}[!t] 
	\centering
	\includegraphics[width=3.4in]{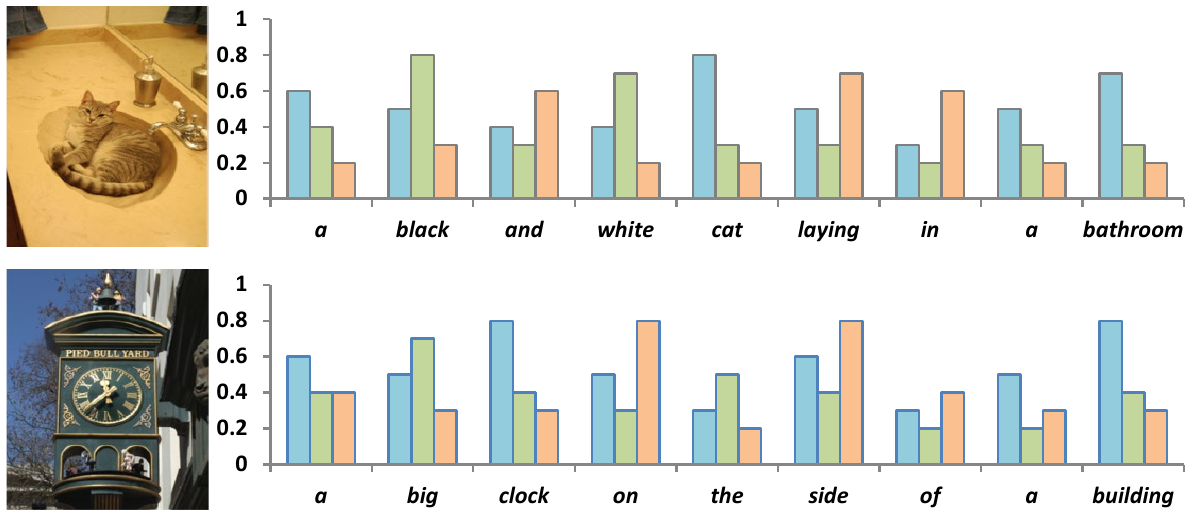}
	\vspace{-0.3cm}
	\caption{
		Visualization of the generated captions, and the per-word gating weights  ($\boldsymbol{\beta}_{t}$) of them belonging to each of the three categories: object, attribute, and relationship.  		
	}
	\vspace{-0.3cm}
	\label{fig:gates}
\end{figure}

\paragraph{Visualization of Gating Weights. } 
To better understand the effect of our context gated fusion attention, we visualize the gating weights ($\boldsymbol{\beta}_t$ in Eqn.\ \ref{eqn:gate}) of object, attribute, and relationship categories 
for each word in the generated captions in Figure~\ref{fig:gates}. 
Our model successfully learns to attend to the category of VSUs 
that are consistent with the type of the current word, \ie object, attribute, or relationship. 
For example, for the verbs like ``laying", the weights for the relationship category are generally the highest. 
The same observations could be found for the adjectives like ``black" and ``big", and the nouns like ``cat" and ``clock".

\begin{figure*}[!tp] 
	\centering
	\includegraphics[width=7in]{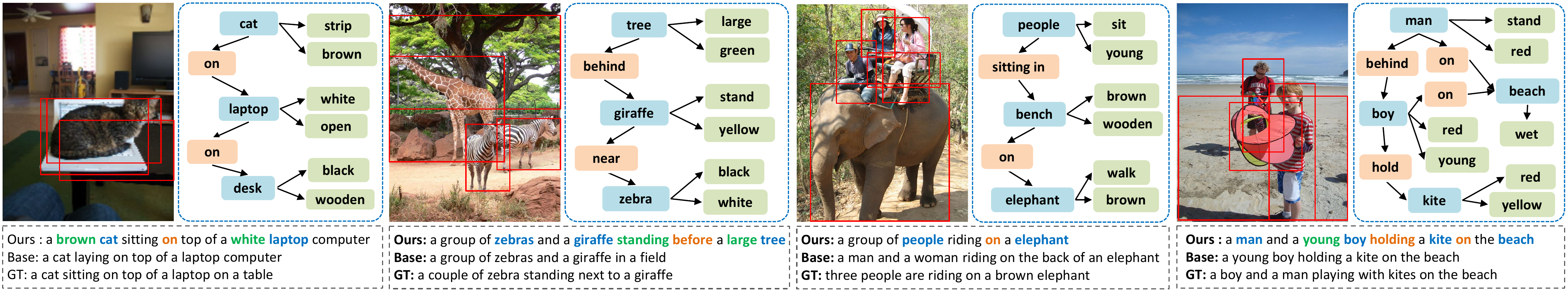}
	\vspace{-0.6cm}
	\caption{
		Example results of the generated captions (by our model, Up-Down baseline, and ground truth) and semantic graphs. 
		Objects, attributes, and relationships are colored with blue, green, and orange, respectively. 
	}
	\vspace{-0.3cm}
	\label{fig:examples}
\end{figure*}

\paragraph{Example Results. }
Figure~\ref{fig:examples} shows four examples of the generated captions and semantic graphs for the images, 
where ``ours", ``base", ``GT" denotes captions from our $\mathcal{G}^S+gate$ model, the Up-Down baseline, and the ground-truth, respectively.  
Generally, our model can generate more descriptive captions than Up-Down by 
enriching the sentences with more precise recognition of objects, detailed description of attributes,   
and comprehensive understanding of interactions between objects. 
For instance, in the second image, our model generates ``standing before a large tree", 
depicting the image content more comprehensively, while the base model fails to recognize 
the tree and the interactions between the animals and the tree. 

\section{Conclusion}
We proposed to fill the information gap between visual content and linguistic description with visual semantic units (VSUs), which are visual components about objects, their attributes, and object-object interactions. 
We leverage structured graph (both semantic graph and geometry graph) to uniformly represent and GCNs to contextually embed the VSUs. 
A novel context gated attention module is introduced to hierarchically align words and VSUs. 
Extensive experiments on MS COCO have shown the superiority of our method.

\bibliographystyle{ACM-Reference-Format}
\bibliography{references}

\end{document}